\documentclass[letterpaper, 10 pt, conference]{ieeeconf}  

\IEEEoverridecommandlockouts                              

\usepackage{amsmath,amsfonts}
\usepackage{amssymb}
\usepackage{dsfont}
\usepackage{bbm}
\usepackage{xcolor}
\usepackage{algorithmic}
\usepackage{algorithm}
\usepackage{booktabs} 
\usepackage{array}
\usepackage[caption=false,font=normalsize,labelfont=sf,textfont=sf]{subfig}
\usepackage{textcomp}
\usepackage{stfloats}
\usepackage{url}
\usepackage{verbatim}
\usepackage{graphicx}
\usepackage{cite}
\usepackage{lipsum}  

\graphicspath{ {./figures/} }

\title{\LARGE \bf Autonomous Algorithm for Training \\ Autonomous Vehicles with Minimal Human Intervention}

\author{Sang-Hyun Lee$^{1 \, *}$, Daehyeok Kwon$^{2,3 \, *}$, and Seung-Woo Seo$^{2}$
\thanks{$^{*} $Equal contribution $^{1}$Sang-Hyun Lee is with the Department of Mobility Engineering, Ajou University, Gyeonggi-do, South Korea        
        {\tt\small (e-mail: sanghyunlee@ajou.ac.kr)}}%
\thanks{$^{2}$Daehyeok Kwon and Seung-Woo Seo are with the Department of Electrical and Computer Engineering, Seoul National University, Seoul, South Korea        
        {\tt\small (e-mail: \{rnzchy03, sseo\}@snu.ac.kr)}}%
\thanks{$^{3}$Daehyeok Kwon is with the ThorDrive Co., Ltd., Seoul, South Korea        
        {\tt\small (e-mail: dhkwon@thordrive.ai)}}%
\thanks{This work has been submitted to the IEEE for possible publication. Copyright may be transferred without notice, after which this version may no longer be accessible.}%
}

\begin{document}

\maketitle
\thispagestyle{empty}
\pagestyle{empty}

\begin{abstract}
Recent reinforcement learning (RL) algorithms have demonstrated impressive results in simulated driving environments. However, autonomous vehicles trained in simulation often struggle to work well in the real world due to the fidelity gap between simulated and real-world environments. While directly training real-world autonomous vehicles with RL algorithms is a promising approach to bypass the fidelity gap problem, it presents several challenges. One critical yet often overlooked challenge is the need to reset a driving environment between every episode. This reset process demands significant human intervention, leading to poor training efficiency in the real world. In this paper, we introduce a novel autonomous algorithm that enables off-the-shelf RL algorithms to train autonomous vehicles with minimal human intervention. Our algorithm reduces unnecessary human intervention by aborting episodes to prevent unsafe states and identifying informative initial states for subsequent episodes. The key idea behind identifying informative initial states is to estimate the expected amount of information that can be obtained from under-explored but reachable states. Our algorithm also revisits rule-based autonomous driving algorithms and highlights their benefits in safely returning an autonomous vehicle to initial states. To evaluate how much human intervention is required during training, we implement challenging urban driving tasks that require an autonomous vehicle to reset to initial states on its own. The experimental results show that our autonomous algorithm is task-agnostic and achieves competitive driving performance with much less human intervention than baselines.
\end{abstract}

\begin{keywords}
Reinforcement learning, deep learning methods, autonomous agents.
\end{keywords}

\section{INTRODUCTION}
\label{section:introduction}
Autonomous driving has been actively researched for decades. The DARPA Urban Challenge \cite{buehler2009darpa}, a milestone event held in 2007, showed that several autonomous vehicles could complete a 60-mile route driving task in a restricted environment \cite{urmson2008autonomous, leonard2008perception}. These achievements spurred further research toward scaling autonomous vehicles to urban driving environments \cite{nothdurft2011stadtpilot, geiger2012team, ziegler2014making, broggi2015proud}. However, these conventional autonomous vehicles struggle to perform human-like behaviors in urban environments, as they are built on rule-based algorithms. To fully realize the massive potential of autonomous vehicles, we must overcome this limitation.

Reinforcement learning (RL) is a promising approach for developing autonomous vehicles that can perform human-like behaviors. Many RL algorithms have achieved impressive results in simulated driving environments. However, deploying autonomous vehicles trained in simulation to the real world remains an open problem due to the fidelity gap between simulated and real-world driving environments. To overcome this gap, several recent works have explored directly training agents with RL algorithms in real-world settings \cite{levine2018learning, nagabandi2020deep, zhu2019dexterous, zeng2020tossingbot}. Kendall et al. \cite{kendall2019learning} demonstrated that off-the-shelf RL algorithms can enable real-world autonomous vehicles to learn driving strategies for lane following. However, their appealing experimental results reveal additional challenges that must be addressed.

One critical but often neglected challenge is the need to reset an environment after each episode. Most RL algorithms assume such repetitive resets to provide multiple attempts and to reduce experience bias. While resetting environments is straightforward in simulated settings, it involves substantial human intervention in the real world \cite{eysenbach2018leave, zhu2019ingredients, lee2024self}. Imagine training a real-world autonomous vehicle to solve roundabout scenarios. We must determine when to abort an episode to prevent the autonomous vehicle from entering unsafe states, such as collisions with surrounding objects. After aborting the episode, we must identify an initial state that allows the autonomous vehicle to collect informative transitions in the subsequent episode, and then manually drive it to the identified initial state. Since these interventions can lead to poor training efficiency, minimizing them is essential for training real-world autonomous vehicles.

In this paper, we propose a new and general autonomous algorithm that enables off-the-shelf RL algorithms to train autonomous vehicles with minimal human intervention. Our algorithm reduces unnecessary human intervention by aborting episodes to prevent unsafe states and identifying informative initial states for subsequent episodes. The key idea behind identifying informative initial states is to estimate how informative an initial state is based on the expected amount of information obtainable from under-explored yet reachable states. Interestingly, our autonomous algorithm can easily allow an autonomous vehicle to collect transitions from such under-explored yet reachable states where it has not been trained.

Our autonomous algorithm takes advantage of rule-based autonomous driving algorithms to return autonomous vehicles to initial states for subsequent episodes. Autonomous vehicles must be returned by safely handling diverse driving tasks in a reset route. This poses a significant challenge for previous autonomous algorithms that assume a single reset task or depend on randomized reset behaviors. Leveraging rule-based algorithms in our work is inspired by prior experimental results, demonstrating that while the rule-based algorithms cannot infer human-like behaviors, they can perform safe and rule-abiding behaviors in diverse driving scenarios \cite{nothdurft2011stadtpilot, geiger2012team, ziegler2014making, broggi2015proud}. In contrast to most recent RL algorithms that overlook the benefits of the rule-based algorithms \cite{kendall2019learning, isele2018navigating, toromanoff2020end, chen2021interpretable}, our algorithm revisits these benefits and leverages them to reduce human intervention.

The main contribution of our work is an autonomous algorithm that enables off-the-shelf RL algorithms to train autonomous vehicles with minimal human intervention. To the best of our knowledge, our work is the first to propose an autonomous algorithm for training autonomous vehicles in the real world. Our autonomous algorithm is applicable to diverse driving scenarios and compatible with any RL algorithm. Furthermore, our work sheds new light on the benefits of rule-based algorithms in reducing human intervention. To evaluate how much human intervention is required during training, we introduce challenging urban driving tasks that require an autonomous vehicle to return to initial states by itself. The experimental results demonstrate that our autonomous algorithm enables autonomous vehicles to learn safe and interactive behaviors in these tasks with significantly less human intervention than baselines.

\section{RELATED WORKS}
\label{section:related works}
A lot of works have demonstrated that autonomous vehicles can perform safety-aware behaviors following traffic rules \cite{nothdurft2011stadtpilot, geiger2012team,  ziegler2014making, broggi2015proud}. Nothdurft et al. \cite{nothdurft2011stadtpilot} introduced one of the first autonomous vehicles that performed successful test drives in real-world urban environments. Their autonomous vehicle handled several urban driving scenarios in Braunschweig’s inner city ring road. Broggi et al. \cite{broggi2015proud} conducted the challenging autonomous driving test that deployed their autonomous vehicle on open public roads in Parma, including diverse intersections and roundabouts. The experimental results of these works are obviously impressive. However, their algorithms rely on task-specific rules or constraints, which limit their scalability. Most recent works have focused on replacing these rule-based algorithms to address this limitation. In contrast, our work shows that beyond merely replacing them, these algorithms can be utilized to reduce human intervention in training real-world autonomous vehicles. 

\begin{figure*}[t]
\vskip 0.1in
\begin{center}
\centerline{\includegraphics[width=0.9\textwidth, trim=8 8 8 8, clip]{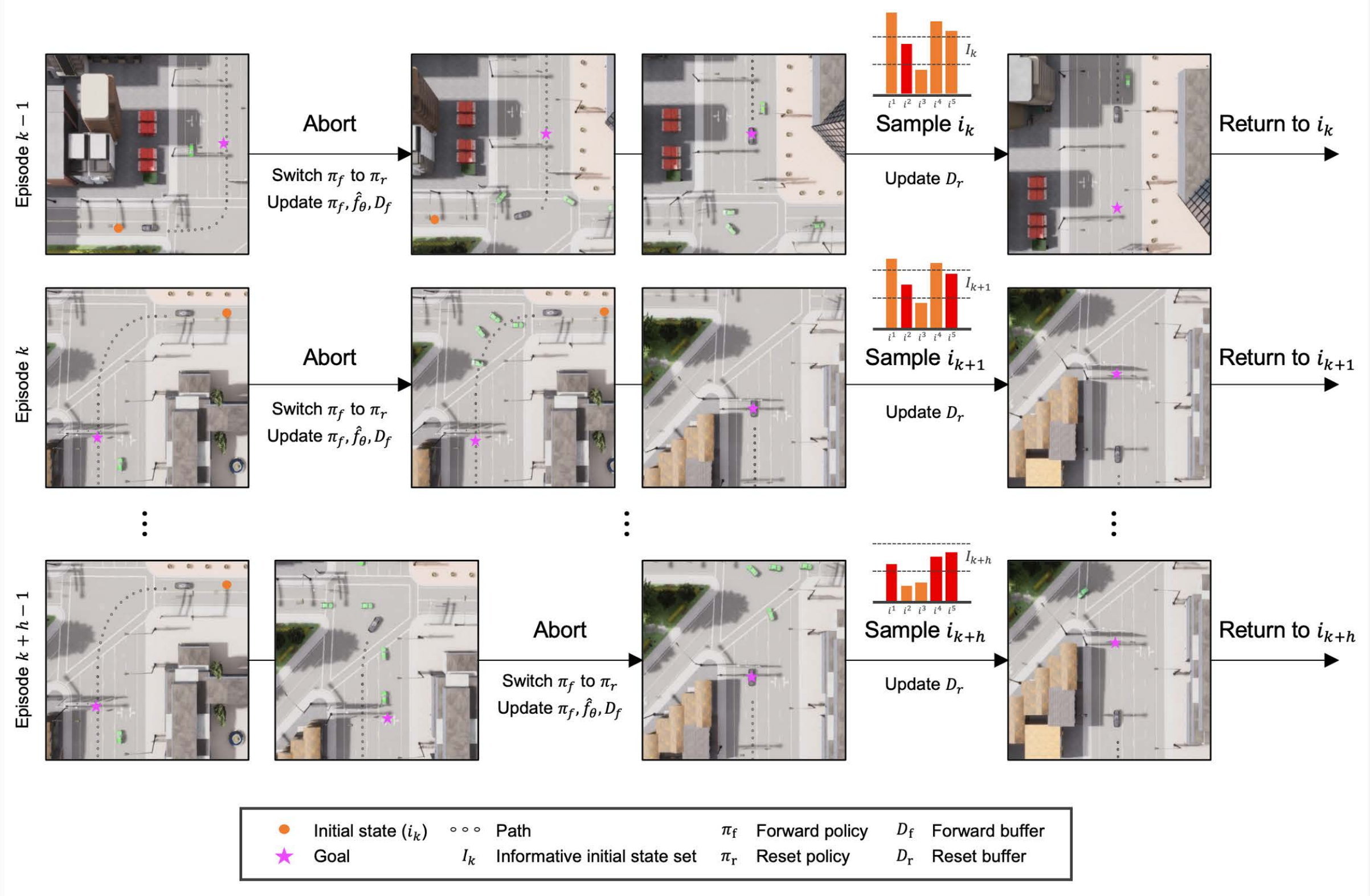}}
\vskip -0.10in
\caption{
Overview of our autonomous algorithm. Our algorithm aborts an episode when the estimated novelty of the current state is too high. After that, an autonomous vehicle is controlled with the reset policy to return to the next initial state without human intervention. The next initial state is sampled from a set of informative initial states. The timing of the switch to the reset policy is pushed further back as the training progresses.
}
\label{fig:overview}
\end{center}
\vskip -0.15in
\end{figure*}

Several recent works on RL have demonstrated that autonomous vehicles can be trained to handle diverse driving scenarios without rule-based algorithms \cite{isele2018navigating, toromanoff2020end, chen2021interpretable}. Isele et al. \cite{isele2018navigating} investigated the effectiveness of RL in handling unsignalized and occluded intersections. Their experimental results demonstrate that the autonomous vehicle trained with an RL algorithm understands diverse occluded intersection scenarios and achieves more robust and efficient performance than rule-based algorithms. Toromanoff et al. \cite{toromanoff2020end} introduced an end-to-end RL algorithm that enables autonomous vehicles to follow traffic lights and avoid surrounding objects. They use an encoder to extract semantic features from raw images and then take the extracted features as the input of their RL algorithm. While these works introduced promising RL algorithms for autonomous vehicles, they assumed that resetting driving environments between every episode is handled outside the training procedure. Kendall et al. \cite{kendall2019learning} showed that resetting driving environments requires substantial human intervention in the real world, leading to poor training efficiency. Our autonomous algorithm can overcome this challenge by resetting driving environments with minimal human intervention.

Our work is inspired by several impressive autonomous algorithms in diverse domains \cite{eysenbach2018leave, zhu2019ingredients, lee2024self}. Eysenbach et al. \cite{eysenbach2018leave} proposed an autonomous algorithm called LNT that induces a curriculum by aborting an episode based on a reset value function. The reset value function is trained with pre-defined reset reward functions. This work empirically shows that the reset value function can be used to prevent an agent from entering irreversible states. However, while autonomous driving tasks require reset behaviors that can address multitask settings, this work assumes that reset behaviors are trained in single-task settings. Zhu et al. \cite{zhu2019ingredients} and Lee et al. \cite{lee2024self} introduced autonomous algorithms that encourage reset policies to continuously discover diverse and unseen initial states. These works implement the reset policies with an exploration algorithm called RND \cite{burda2018exploration}. While these algorithms are scalable and task-agnostic, applying them to train autonomous vehicles is not straightforward, as their reset policies may not guarantee safety-aware behaviors. To the best of our knowledge, no previous work has introduced an autonomous algorithm designed for training autonomous vehicles. Our autonomous algorithm aims to address this gap by exploring how to train autonomous vehicles to handle diverse driving scenarios while reducing human intervention.

\section{TRAINING AUTONOMOUS VEHICLES WITH \\ MINIMAL HUMAN INTERVENTION}
In this section, we introduce an autonomous algorithm that enables off-the-shelf RL algorithms to train autonomous vehicles with minimal human intervention. The goal of our autonomous algorithm is to empower autonomous vehicles with the ability to continuously learn and improve by themselves. To achieve this goal, our autonomous algorithm learns both when to abort an episode to prevent autonomous vehicles from entering unsafe states and where to return them to collect informative transitions in the following episodes. Figure 1 provides an overview of the training procedure of our algorithm.

\subsection{Problem Formulation}
We use a Markov decision process (MDP) to model an environment. MDP is defined as the tuple $(S, A, P, R, \rho_0, \gamma)$, where $S$ denotes the set of states, $A$ denotes the set of actions, and $P: S \times A \times S \rightarrow \mathbb{R}^+$ denotes the state transition model. The function $R: S \times A \times S \rightarrow \mathbb{R}$ denotes the reward function, which outputs a scalar feedback called a reward, $r$. $\rho_0: S \rightarrow \mathbb{R}^+$ represents the initial state distribution, and $\gamma$ represents the discount factor. Key components that represent behaviors of an agent are the forward policy and the forward state-action value function: The forward policy $\pi_f(a|s)$ maps a state to a probability distribution over actions and the forward state-action value function $Q^{\pi_f}(s, a)$ represents the expected return obtained when the agent takes the action $a$ in the state $s$ and follows the policy $\pi_f$.

RL aims to find the optimal policy $\pi_f^*$ that maximizes the expected cumulative rewards when the state transition model is unknown. While RL algorithms have achieved remarkable results in various domains, they typically assume that resetting environments is managed outside the training procedure \cite{mnih2015human, vinyals2019grandmaster}. This assumption makes it difficult to apply such algorithms to train autonomous vehicles in the real world, as resetting driving environments demands significant human intervention. 

Our work seeks to address three main challenges that are critical for minimizing human intervention. First, we must decide when to abort an episode to prevent autonomous vehicles from entering unsafe states, such as collisions with surrounding objects. Resetting from unsafe states requires significant intervention and may even make further training impossible. Second, after aborting an episode, we must safely return autonomous vehicles to an initial state for a subsequent episode while complying with traffic rules. This poses a challenge for most previous autonomous algorithms that cannot perform safety-aware reset behaviors. Finally, we must identify which initial state can provide autonomous vehicles with informative transitions in a subsequent episode. Initial states that are too easy or too difficult can preclude autonomous vehicles from obtaining informative transitions or can easily lead them into irreversible states. In the remainder of this section, we discuss how our autonomous algorithm addresses each of these challenges in detail.

\subsection{Aborting Episodes to Prevent Unsafe States}
\label{sec:preventing_irreversible}
One of the main difficulties in preventing an autonomous vehicle from entering unsafe states is assessing the safety of a state. There are no dominant or widely accepted approaches.  To assess state safety, we must take into account the learning progress of an autonomous vehicle. Furthermore, state safety must be computationally tractable in real time. Our autonomous algorithm addresses this difficulty by leveraging the concept of novelty to approximate state safety. Specifically, our algorithm estimates the novelty of the current state at each time step and then aborts an episode if the estimated novelty is too high. This is based on the hypothesis that autonomous vehicles may not know how to avoid unsafe states on under-explored state space, where the state novelty is typically high. As shown in Section \ref{sec:experimental_results_and_analysis}, we empirically observed that the estimated state novelty is high when an autonomous vehicle being trained enters unsafe states.

Fortunately, several recent works have introduced feasible state-novelty estimation approaches \cite{bellemare2016unifying, pathak2017curiosity, burda2018exploration}. In our experiments, we used random network distillation (RND) \cite{burda2018exploration} to estimate the novelty of states, as it is simple to implement and works well in high-dimensional observations. RND defines a target network $f(s): S \rightarrow \mathbb{R}^k$ and a predictor network $\hat{f}_\theta(s): S \rightarrow \mathbb{R}^k$. The target network is randomly initialized and then fixed, and the predictor network is trained to minimize the expected prediction error $\|\hat{f}_\theta(s)-f(s)\|^2$ with collected transitions. RND empirically demonstrated that the prediction error is higher in unseen states than in frequently explored states. Based on this interesting experimental result, we regard the prediction error as the estimated novelty of a state. Note that aborting episodes based on state novelty is task-agnostic. It contrasts with several previous works that require task-specific knowledge, such as reset reward functions or demonstrations, to abort episodes \cite{eysenbach2018leave,sharma2021autonomous}.

\subsection{Returning with Safety-aware Reset Behaviors}
After aborting an episode, an autonomous vehicle must return to an initial state for the subsequent episode. The reset process requires the autonomous vehicle to perform safety-aware behaviors and adhere to traffic rules. Furthermore, the autonomous vehicle must handle a sequence of diverse driving tasks during the reset process. Imagine training an autonomous vehicle to address a detour task. The route from the aborted state to the initial state for the subsequent episode is likely to include other driving tasks, such as lane-change and intersection tasks. These challenges make it difficult to apply previous autonomous algorithms to train autonomous vehicles, as these algorithms depend on randomized reset behaviors or assume the reset process involves a single task. 

While we learn both when to abort an episode and where to return autonomous vehicles for subsequent episodes, we leverage rule-based autonomous driving algorithms to return autonomous vehicles with safety-aware and rule-abiding behaviors. This is based on prior experimental results as follows: rule-based algorithms cannot perform flexible behaviors like expert drivers, but they can perform safety-aware and rule-abiding behaviors across diverse driving scenarios. Most RL algorithms overlook the benefits of rule-based algorithms. In contrast, our work sheds new light on these benefits for training autonomous vehicles in the real world. Note that our autonomous algorithm is compatible with any rule-based autonomous driving algorithm.

\subsection{Identifying Informative Initial States}
Initial states determine the transitions an autonomous vehicle encounters during training. Some initial states can guide an autonomous vehicle to under-explored states, where it might choose unsafe actions. This causes humans to intervene and abort an episode early. The early abort hinders collecting sufficient informative transitions. Conversely, other initial states can guide an autonomous vehicle to too familiar states, where it rarely collects informative transitions even if episodes are completed without aborts. Identifying informative initial states is therefore critical for enabling an autonomous vehicle to collect sufficient informative transitions without human intervention. Since such initial states are not typically given in the field of autonomous driving, an autonomous vehicle must identify them on its own.

Our autonomous algorithm enables an autonomous vehicle to identify informative initial states that are neither too under-explored nor too familiar. The key idea is to assess how informative an initial state is based on the expected amount of information obtainable from reachable but under-explored states. The initial state for the $k$th episode, $i_k$, can then be determined as follows:
\begin{equation} \label{eq:informative_initial_state}
i_k \sim  \text{Unif}(I_k), \:\: \text{where} \:\: I_k \triangleq \{i \in I \: | \: \lambda_1 \leq e_i \leq \lambda_2 \}.
\end{equation}
$I_k$ is the set of informative initial states for the $k$th episode, $i$ is an initial state included in the set of all initial states $I$, $e_i$ is the expected amount of obtainable information, and $\lambda_1$ and $\lambda_2$ are its lower and upper bounds, respectively. The initial state for the subsequent episode is uniformly sampled from the set of informative initial states identified at the current episode. This can prevent an autonomous vehicle from resetting to either too under-explored or too familiar initial states. We would like to emphasize that the set of informative initial states can adapt to the learning progress of an autonomous vehicle, suggesting that our autonomous algorithm implicitly generates a curriculum for initial states. Experimental results described in Section \ref{sec:experimental_results_and_analysis} indicate that identifying informative initial states can reduce human intervention by achieving better sample efficiency.

The under-explored states reachable from an initial state are those that an autonomous vehicle encounters between aborting an episode and reaching a goal state. Our algorithm easily enables an autonomous vehicle to collect transitions from the under-explored but reachable states by returning it to the initial state via the goal state after an episode is aborted. Using the transitions collected from these states, our algorithm estimates the expected amount of obtainable information by evaluating their novelty based on RND prediction errors, similar to the state novelty estimation discussed in Section \ref{sec:preventing_irreversible}. The expected amount of obtainable information for an initial state $i$ can then be written as follows:
\begin{equation} \label{eq:informative_initial_state_set}
e_i = \mathbb{E}_{s \sim D_r^i}\big[\|\hat{f}_{\theta}(s) - f(s) \|\big],
\end{equation}
where $D_r^i$ is the reset buffer for an initial state $i$. Note that the transitions sampled from the reset buffer are not used to train the predictor network $\hat{f}_\theta(s)$. The predictor network is trained with the transitions collected before aborting episodes.

\subsection{Training Procedure Details}
Algorithm \ref{alg:training} describes the overall training procedure of our autonomous algorithm. Before aborting an episode, an autonomous vehicle is trained and controlled with the RL forward policy $\pi_f(a|s)$. After the episode is aborted, the vehicle is controlled with the rule-based reset policy $\pi_r(a|s)$ to reach a goal state. Once the vehicle reaches the goal state, the informative initial state set for the subsequent episode is identified, and the reset policy returns the vehicle to an initial state sampled from the identified set. To estimate the informative initial state set, we define independent reset buffers for each initial state, which makes it efficient to sample corresponding transitions. We would like to note that the forward policy trained with an RL algorithm can outperform the rule-based reset policy.

\begin{figure}[t]
\begin{algorithm}[H]
   \caption{Overall Training Procedure}
   \label{alg:training}
\algsetup{linenosize=\small}
\begin{algorithmic}[1]
   \STATE {\bfseries Given:} Initial state set $I_1$
   \STATE Initialize forward policy and buffer $\pi_f(a|s), D_f$
   \vspace{0.05mm}
   \STATE Initialize reset buffers $\{D_r^j\}_{j=1,\dots,n}$
   \vspace{0.05mm}
   \STATE Initialize target and predictor networks $f(s), \hat{f}_\theta(s)$
   \vspace{0.05mm}   
   \STATE Sample initial state $i_1 \sim \text{Unif}(I_1) $
   \vspace{0.05mm}   
   \FOR{$k \leftarrow 1 \dots K$}
   \vspace{0.05mm}
   \FOR{$t \leftarrow 1 \dots T_\text{forward}$}
   \vspace{0.05mm}
   \IF{$\lambda_0 \leq \|\hat{f}_\theta(s_t) - f(s_t)\|$}
   \vspace{0.05mm}
   \STATE Abort and switch to reset policy $\pi_r(a_t|s_t)$
   \vspace{0.05mm}
   \ENDIF
   \vspace{0.05mm}
   \STATE Select forward action $a_t \sim \pi_f(a_t|s_t)$
   \vspace{0.05mm}      
   \STATE Obtain and add transition to forward buffer $D_f$ \\   
   \vspace{0.05mm}
   \STATE Update forward policy $\pi_f(a|s)$ and predictor $\hat{f}_\theta(s)$
   \ENDFOR
   \vspace{0.05mm}

   \FOR{$t \leftarrow 1 \dots T_\text{reset}$}
   \vspace{0.05mm}
   \STATE Select reset action $a_t \sim \pi_r(a_t|s_t)$
   \vspace{0.05mm}
   \STATE Obtain and add transition to reset buffer $D^{i_{k}}_r$ \\   
   \vspace{0.15mm}
   \ENDFOR
   \STATE Estimate informative initial state set $I_{k+1}$\\
   \vspace{0.3mm}
   \STATE Sample next initial state $i_{k+1} \sim \text{Unif}(I_{k+1})$   
   \vspace{0.05mm}      
   \STATE Return to initial state $i_{k+1}$ with reset policy $\pi_r(a|s)$   
   \vspace{0.05mm}
   \ENDFOR
\end{algorithmic}
\end{algorithm}
\vskip -0.15in
\end{figure}

\section{EXPERIMENTS}
Our experiments aim to answer the following questions: 1) Can our autonomous algorithm enable autonomous vehicles to achieve competitive driving performance? 2) Can our autonomous algorithm reduce human intervention required in training autonomous vehicles? and 3) How does identifying informative initial states affect the performance of our autonomous algorithm? To answer these questions, we introduce five urban driving tasks and use them to evaluate our autonomous algorithm against baselines.

\subsection{Baselines}
The baselines used in our experiments are as follows: 1) an agent that randomly selects actions with access to external resets (Random), 2) an RL agent that has access to external resets (Oracle), 3) LNT that is one of the state-of-the-art autonomous algorithms (LNT) \cite{eysenbach2018leave}, and 4) a variant of our autonomous algorithm that leverages rule-based autonomous driving algorithms without episode aborts and informative initial state identification (Ours w/o Curr). LNT relies on predefined reset reward functions to learn when to abort episodes and does not identify informative initial states. In contrast, our algorithm can learn when to abort episodes without reset reward functions and can identify informative initial states. Furthermore, it enables an autonomous vehicle to perform safety-aware reset behaviors. Note that since the other state-of-the-art autonomous algorithms \cite{zhu2019ingredients, lee2024self} mentioned in section \ref{section:related works} perform randomized reset behaviors, we did not use them as baselines in our experiments.

\begin{table}[t]
\vspace{0.1in}
\caption{Hyperparameters}
\begin{center}
\vskip -0.1in
\resizebox{0.98\columnwidth}{!}{
\begin{tabular}{c c}
\toprule
\qquad HYPERPARAMETER \qquad \qquad & \qquad VALUE \qquad \qquad \\
\midrule
\qquad Batch Size \qquad \qquad & \qquad 256 \qquad \qquad \\
\qquad Buffer Size (Forward) \qquad \qquad & \qquad 50000 \qquad \qquad \\
\qquad Buffer Size (Reset) \qquad \qquad & \qquad 1000 \qquad \qquad \\
\qquad Learning Rate \qquad \qquad & \qquad $1\times10^{-3}$ \qquad \qquad \\ 
\qquad Discount Factor \qquad \qquad & \qquad 0.99 \qquad \qquad \\
\qquad Temperature \qquad \qquad & \qquad 0.4 \qquad \qquad \\
\qquad Gradient Step \qquad \qquad & \qquad 1 \qquad \qquad \\
\qquad Target Update Interval \qquad \qquad & \qquad 1 \qquad \qquad \\ 
\qquad Target Smoothing Coefficient \qquad \qquad & \qquad 0.005 \qquad \qquad \\
\qquad $\lambda_0 $ \qquad \qquad & \qquad 1.4 \qquad \qquad \\  
\qquad $\lambda_1$ \qquad \qquad & \qquad 1.0 \qquad \qquad \\  
\qquad $\lambda_2$ \qquad \qquad & \qquad 1.7 \qquad \qquad \\  
\bottomrule
\end{tabular}
}
\label{tab:hyperparameters}
\end{center}
\vskip -0.1in
\end{table}

\subsection{Implementation Details}
\label{sec:implementation_details}
Our autonomous algorithm includes two main learnable models: the forward policy and the RND predictor, each represented by neural networks. The forward policy has two hidden layers of 512 units with ReLU activations and an additional softmax layer to output a categorical distribution over high-level actions, such as go, crawl, and stop. The input of the forward policy consists of the predicted trajectories of surrounding vehicles and the planned trajectories of an autonomous vehicle. We used Soft Actor-Critic (SAC) \cite{haarnoja2018soft}, which is a state-of-the-art off-policy RL algorithm, and the Adam optimizer to update the forward policy. To provide a fair comparison, the forward policies of our algorithm and baselines were designed to have the same structure and were trained with the same RL algorithm and optimizer. 

The RND predictor has three convolutional layers and a linear output layer as follows: 32 filters of size 8×8 and stride 4, 64 filters of size 4×4 and stride 4, and 32 filters of size 3×3 and stride 1. The output of the last convolutional layer is fed into a linear layer that has 512 hidden units and outputs 256-dimensional feature vectors. The input of the RND predictor is a bird’s-eye view segmentation mask. Similar to the forward policy, we utilized Adam optimizer to update the RND predictor. The key hyperparameters used in our experiments are described in Table \ref{tab:hyperparameters}, and they were tuned with the coarse grid search. 

\begin{table*}[t]
\vskip 0.1in
\caption{Quantitative Results on Urban Driving Tasks \label{tab:quantitative_results}}
\begin{center}
\vskip -0.1in
\resizebox{1.0\textwidth}{!}{
\begin{tabular}{c ccc | ccc | ccc | ccc | ccc}
\toprule
& \multicolumn{3}{c}{DETOUR} & \multicolumn{3}{c}{THREE-WAY} & \multicolumn{3}{c}{FOUR-WAY} & \multicolumn{3}{c}{FIVE-WAY} & \multicolumn{3}{c}{ROUNDABOUT}\\
\toprule
& \: AS $\downarrow$ & SR $\uparrow$ & MR $\downarrow$ \qquad 
& \: AS $\downarrow$ & SR $\uparrow$ & MR $\downarrow$ \qquad 
& \: AS $\downarrow$ & SR $\uparrow$ & MR $\downarrow$ \qquad 
& \: AS $\downarrow$ & SR $\uparrow$ & MR $\downarrow$ \qquad 
& \: AS $\downarrow$ & SR $\uparrow$ & MR $\downarrow$\\
\midrule
Random \qquad & 947.9 & 0.06 & 600.0 \qquad & 773.7 & 0.48 & 600.0 \qquad & 685.4 & 0.58 & 600.0 \qquad & 725.1 & 0.50 & 600.0 \qquad & 920.6 & 0.45 & 500.0 \\
Oracle \qquad & 146.5 & 1.00 & 600.0 \qquad & 414.0 & 0.86 & 600.0 \qquad & 348.8 & 0.87 & 600.0 \qquad & 288.5 & 0.94 & 600.0 \qquad & 453.2 & 0.96 & 500.0 \\
LNT \qquad & 150.0 & 1.00 & 176.0 \qquad & 465.4 & 0.72 & 398.0 \qquad & 493.3 & 0.72 & 570.0 \qquad & 471.4 & 0.74 & 538.0 \qquad & 516.4 & 0.86 & 312.0 \\
Ours (w/o Curr) \qquad & 151.5 & 1.00 & 108.7 \qquad & 408.4 & 0.81 & 129.0 \qquad & 351.9 & 0.84 & 144.0 \qquad & 290.1 & 0.93 & \: 75.0 \qquad & 492.0 & 0.92 & \: 73.0 \\
\textbf{Ours} \qquad & \textbf{151.5} & \textbf{1.00} & \: \textbf{70.1} \qquad & \textbf{404.3} & \textbf{0.85} & \: \textbf{85.0} \qquad & \textbf{347.8} & \textbf{0.87} & \textbf{104.0} \qquad & \textbf{281.6} & \textbf{0.94} & \: \textbf{54.0} \qquad & \textbf{476.0} & \textbf{0.95} & \: \textbf{47.0}\\
\bottomrule
\end{tabular}
}
\end{center}
\vskip -0.1in
\end{table*}

The reset policy is implemented with the autonomous driving agent provided by the open-source simulator CARLA \cite{dosovitskiy2017carla}. The agent uses reliable rule-based algorithms designed to ensure safety and follow traffic rules. We observed that it performed safety-aware and rule-abiding behaviors that could handle most scenarios in our evaluation tasks. While improving it could also reduce human intervention, we leave this issue for future work, as it is orthogonal to our autonomous algorithm.

\begin{figure*}[t]
\begin{center}
\centerline{\includegraphics[width=1.0\textwidth, trim=8 8 8 8, clip]{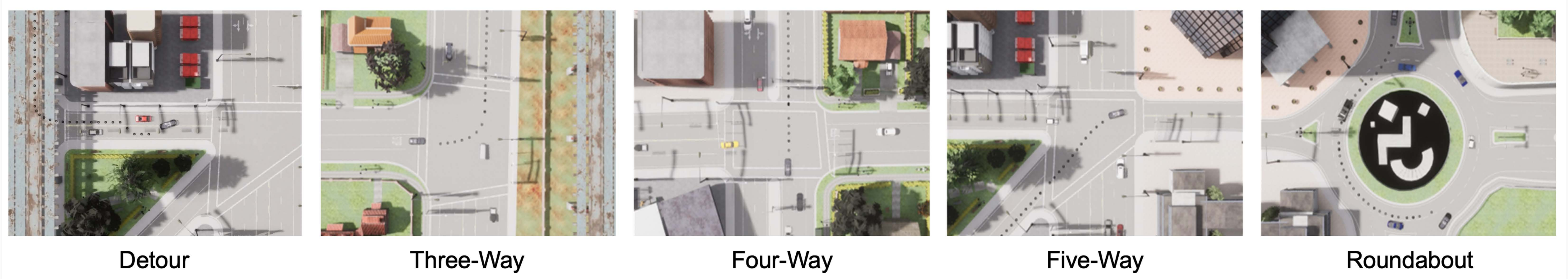}}
\vskip -0.1in
\caption{Urban driving tasks introduced in our experiments. All spawned surrounding vehicles are set to ignore traffic signals, so an autonomous vehicle being trained should consider interactions with them to solve these tasks. The black dotted line denotes the route to a given goal.}
\label{fig:environments}
\end{center}
\vskip -0.2in
\end{figure*}

\subsection{Environments}
Existing benchmarks in the field of autonomous driving focus on evaluating the driving performance of trained autonomous vehicles. These benchmarks do not provide metrics and settings to evaluate autonomous algorithms that aim to reduce human intervention. Therefore, we introduce challenging urban driving tasks that require an autonomous vehicle to reset to initial states on its own, allowing us to measure how much human intervention is required during training. Figure \ref{fig:environments} describes these tasks introduced in our work, which consist of a detour, a roundabout, and three different unsignalized intersection tasks.  Similar to the reset policy discussed in Section \ref{sec:implementation_details}, they were implemented within CARLA \cite{dosovitskiy2017carla}. The goal across these tasks is to reach given target locations as fast as possible without collisions, and an autonomous vehicle must understand interactions with surrounding vehicles to achieve this goal. The reward function for the detour task is defined as follows:
\begin{equation} \label{eq:detour_reward}
\begin{aligned}
r(s_t, a_t) = \lambda_g & \mathds{1}_g(s_t,a_t) - \lambda_p\mathds{1}_p(s_t,a_t) \\
& \qquad \: {}- \lambda_c\mathds{1}_c(s_t,a_t) - \lambda_s\mathds{1}_s(s_t,a_t),
\end{aligned}
\end{equation}
where $\mathds{1}_g$ indicates whether our agent reaches a goal, $\mathds{1}_p$ indicates whether our agent crosses a center line, $\mathds{1}_c$ indicates whether our agent collides or fails to keep a safe distance, $\mathds{1}_s$ indicates whether our agent survives, and $\lambda_g$, $\lambda_p$, $\lambda_c$, and $\lambda_s$ are hyperparameters to balance each of these terms, respectively. The shared reward function for other tasks is defined as follows:
\begin{equation} \label{eq:intersection_reward}
r(s_t, a_t) = \lambda_v \, \frac{v_t}{v_{\text{max}}} - \lambda_c\mathds{1}_c(s_t,a_t) - \lambda_s\mathds{1}_s(s_t,a_t),
\end{equation}
where $v_t$ denotes the current speed, $v_\text{max}$ denotes the maximum speed, and $\lambda_v$ is the corresponding hyperparameter.

\subsection{Experimental Results and Analysis}
\label{sec:experimental_results_and_analysis}
The evaluation metrics used in our experiments are as follows: success rate (SR), average episode step (AS), and the number of manual resets (MR). Success indicates whether an autonomous vehicle reaches a goal within a time limit without collision, and the average episode step represents how efficiently an autonomous vehicle reaches a goal. The number of manual resets indicates how much human intervention is required to train an autonomous vehicle. While both SR and AS are calculated in the evaluation procedure, MR is calculated throughout the training procedure. A manual reset is triggered when an autonomous vehicle enters an irreversible state due to a collision, fails to reach a goal, or cannot return to an initial state for the subsequent episode. 

Table \ref{tab:quantitative_results} describes the numerical training results computed over 100 episodes for each urban driving task. While Oracle achieves the best performance across all driving tasks, it requires repetitive manual resets after every episode. This makes it impractical to use Oracle for training autonomous vehicles in real-world driving environments. LNT triggers a much larger number of manual resets than Ours (w/o Curr). This gap can be attributed to the safety-aware reset behaviors of our autonomous algorithm, which demonstrates the benefits of rule-based algorithms in reducing human intervention. We empirically observed that the reset policy of LNT could not simultaneously learn all tasks encountered in the reset route. Our algorithm achieves competitive driving performance with significantly fewer manual resets than baselines. In particular, the gap between Ours and Ours (w/o Curr) implies that learning when to abort episodes and identifying informative initial states are critical in reducing unnecessary manual resets.

\begin{figure*}[t]
\vskip 0.1in
\begin{center}
\centerline{\includegraphics[width=0.85\textwidth, trim=8 8 8 8, clip]{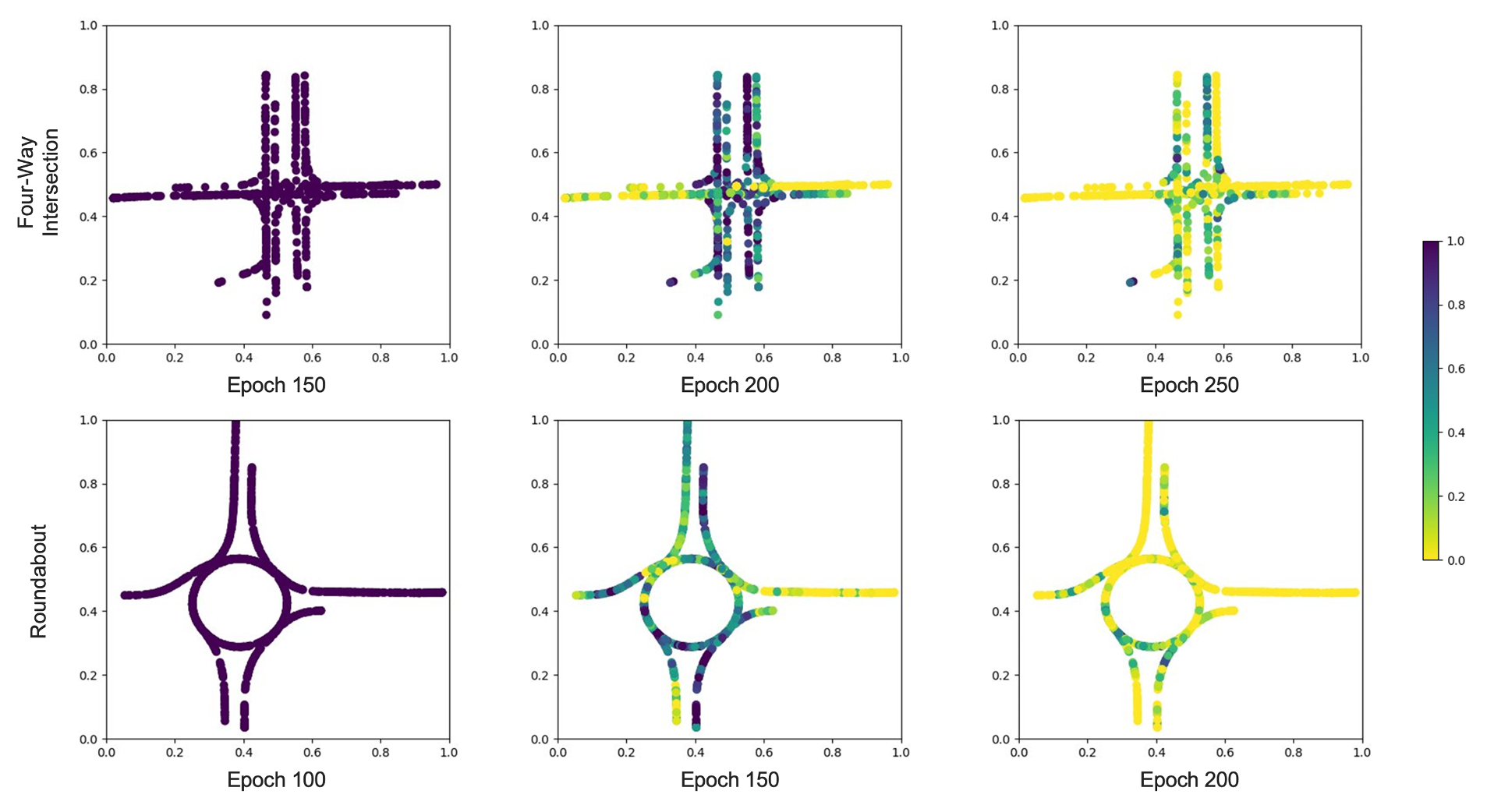}}
\caption{Estimated novelty of sampled states in four-way unsignalized intersection and roundabout tasks. Each dimension of the states is normalized to [0, 1], and their colors represent the novelty estimated by the RND predictor. These results indicate that the state space where an autonomous vehicle can be trained continually broadens as the training progresses.}
\label{fig:qualitative_pred_errors}
\end{center}
\vskip -0.2in
\end{figure*}

\begin{figure}[t]
\vskip 0.05in
\begin{center}
\centerline{\includegraphics[width=0.85\columnwidth, trim=8 8 8 8, clip]{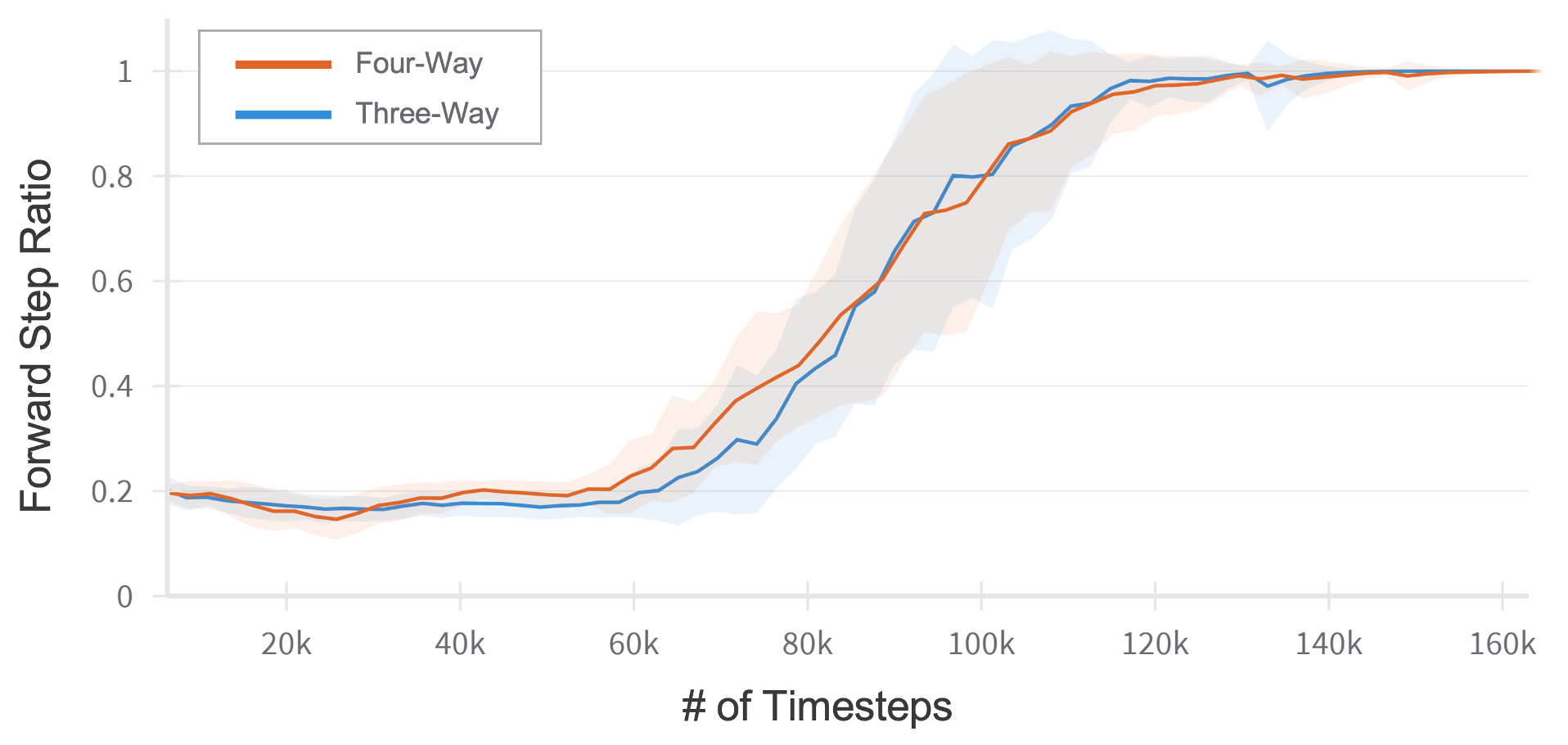}}
\caption{Forward step ratios for three-way and four-way unsignalized intersection tasks. The forward step ratio refers to the ratio between the number of forward time steps and the total number of time steps. The darker-colored lines and shaded areas denote the means and standard deviations over 10 random seeds, respectively.}
\label{fig:forward_reset_ratio}
\end{center}
\vskip -0.2in
\end{figure}

To analyze how our autonomous algorithm prevents autonomous vehicles from entering unsafe states, we calculate the ratio of the number of time steps before an episode is aborted to the total number of time steps. Figure \ref{fig:forward_reset_ratio} describes how this ratio changes over training time on the three-way and four-way unsignalized intersection tasks. We observed that episodes are aborted near initial states in the early stages of training and near goal states at the end of the training. This suggests that our autonomous algorithm can reduce manual resets by preventing an autonomous vehicle from entering under-explored states, where it might take unsafe actions. Note that our algorithm takes into account the learning progress of an autonomous vehicle to determine when to abort episodes.

We also visualize how the states in which our algorithm allows an autonomous vehicle to explore change over time. As mentioned in Algorithm \ref{alg:training}, the estimated novelty of these states should be lower than the abort threshold $\lambda_0$. To visualize such states, we randomly sample some states from multiple rollouts of each initial state and estimate the novelty of the sampled states. Figure \ref{fig:qualitative_pred_errors} illustrates the visualization results in the four-way unsignalized intersection and roundabout tasks. Note that the estimated state novelty is clipped to the range $[ \lambda_0, \lambda_0 + 2]$, and then normalized to [0, 1]. The visualization results show that the states having the estimated novelty below $\lambda_0$ are gradually spread out from initial states as the training progresses. Therefore, Figure \ref{fig:qualitative_pred_errors} can then be interpreted as the qualitative evidence supporting the quantitative results shown in Figure \ref{fig:forward_reset_ratio}.

\begin{figure}[t]
\begin{center}
\centerline{\includegraphics[width=0.85\columnwidth, trim=8 8 8 8, clip]{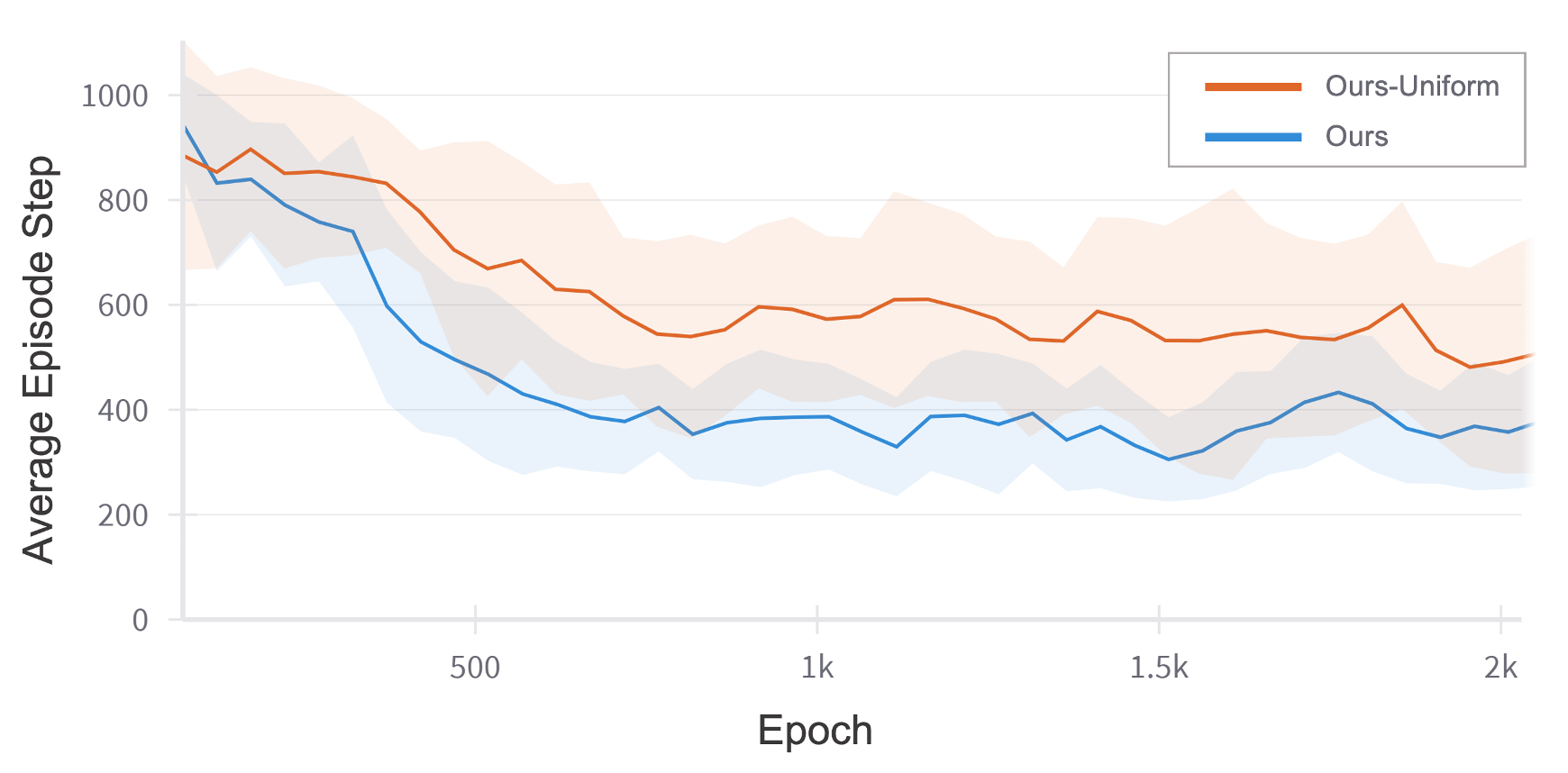}}
\caption{Effects of identifying informative initial states on performance in five-way unsignalized intersection task. Our algorithm attains a lower average episode step and converges faster than the variant that samples initial states uniformly. The darker-colored lines and shaded areas denote the means and standard deviations over 10 random seeds, respectively.}
\label{fig:ablation_step}
\end{center}
\vskip -0.25in
\end{figure}

\begin{figure}[t]
\vskip 0.1in
\begin{center}
\centerline{\includegraphics[width=0.90\columnwidth, trim=8 8 8 8, clip]{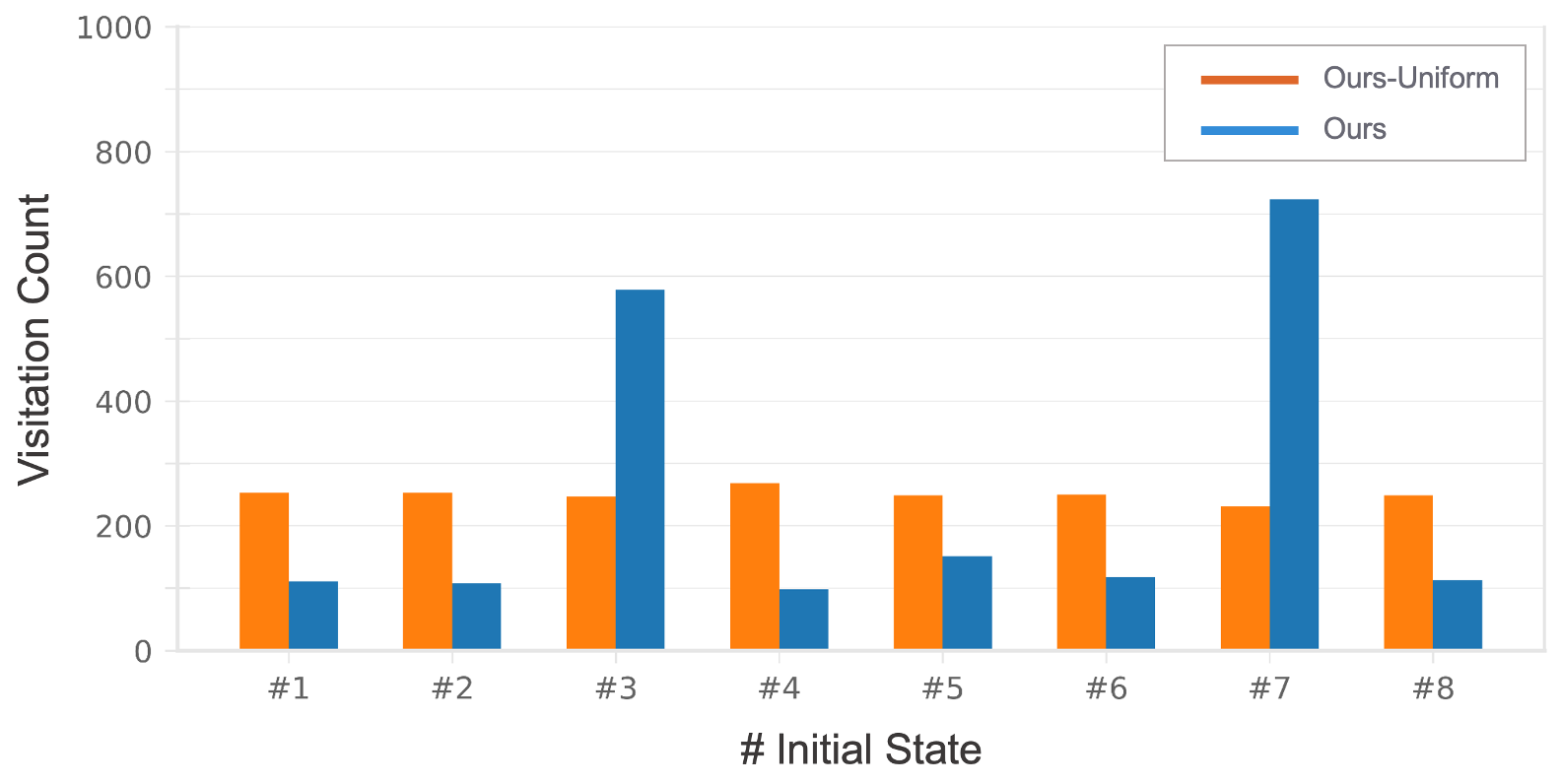}}
\caption{Effects of identifying informative initial states on initial state visitation in five-way unsignalized intersection task. The initial states numbered 3 and 7 are informative, while the others are non-informative.}
\label{fig:ablation_visitation}
\end{center}
\vskip -0.2in
\end{figure}

To better understand the benefits of identifying informative initial states, we ran an ablation study on the five-way unsignalized intersection task, where some initial states are intentionally designed to be non-informative. When an autonomous vehicle resets to these non-informative initial states, surrounding vehicles are not spawned, and the inputs of the RND predictor are randomly shuffled. An autonomous vehicle would struggle to collect informative transitions under these conditions. Figure \ref{fig:ablation_step} describes the performance comparison between our autonomous algorithm and the variant that uniformly samples initial states. We observed that our algorithm achieves a lower average episode step and converges faster than the variant. This indicates that identifying informative initial states contributes to better asymptotic performance and sample efficiency of our algorithm, leading to fewer manual resets. In addition, as shown in Figure \ref{fig:ablation_visitation}, we also confirmed that our algorithm returns an autonomous vehicle to informative initial states much more frequently than non-informative initial states.

\section{CONCLUSION}
We introduce an autonomous algorithm that enables off-the-shelf RL algorithms to train autonomous vehicles with minimal human intervention. The three key challenges addressed in our work are: 1) when to abort episodes to prevent autonomous vehicles from entering unsafe states, 2) where to return autonomous vehicles to collect informative transitions, and 3) how to safely return autonomous vehicles to initial states for subsequent episodes. Experimental results demonstrate that our autonomous algorithm enables autonomous vehicles to achieve competitive driving performance with far fewer manual resets than baselines in diverse urban driving tasks. We will explore the following research directions in future work. First, we will investigate the limitations of using novelty to abort episodes, such as delaying exploration. We expect that uncertainty-based approaches can alleviate the limitations by improving the stability and interpretability of our autonomous algorithm. Second, we will combine our algorithm with offline learning algorithms to accelerate the early stages of training. We believe both algorithms can complement each other in realizing autonomous vehicles in the real world. Finally, we will scale our algorithm to multitask settings. Integrating goal-conditioned RL into our algorithm is a promising approach for this line of research.

\bibliographystyle{./IEEEtran}
\bibliography{./IEEEabrv,./IEEEexample, ./references}

\end{document}